\pdfoutput=1

\documentclass[11pt]{article}

\usepackage{acl}

\newif\ifcomment\commenttrue

\usepackage[a-1b]{pdfx}

\usepackage{framed}
\usepackage{mdwlist}
\usepackage{siunitx}
\usepackage{latexsym}
\usepackage{colortbl}
\usepackage{xcolor}
\usepackage{nicefrac}
\usepackage{booktabs}
\usepackage{fnpct}
\usepackage{amsfonts}
\usepackage[T1]{fontenc}
\usepackage{bold-extra}
\usepackage{amsmath}
\usepackage{amssymb}
\usepackage{bm}
\usepackage{graphicx}
\usepackage{mathtools}
\usepackage{microtype}
\usepackage{multirow}
\usepackage{multicol}
\usepackage{xpatch}
\usepackage{latexsym,comment}
\usepackage[normalem]{ulem}

\newcommand*{\missingreference}{{\Huge \colorbox{red}{?reference?}}}
\newcommand*{\missingcitation}{{\Huge \colorbox{red}{?citation?}}}

\makeatletter
\xpatchcmd{\@setref}{\bfseries}{\missingreference}{}{}
\def\@citex[#1]#2{\leavevmode
    \let\@citea\@empty
    \@cite{\@for\@citeb:=#2\do
        {\@citea\def\@citea{,\penalty\@m\ }%
            \edef\@citeb{\expandafter\@firstofone\@citeb\@empty}%
            \if@filesw\immediate\write\@auxout{\string\citation{\@citeb}}\fi
            \@ifundefined{b@\@citeb}{\hbox{\reset@font\missingcitation}%
                \G@refundefinedtrue
                \@latex@warning
                {Citation `\@citeb' on page \thepage \space undefined}}%
            {\@cite@ofmt{\csname b@\@citeb\endcsname}}}}{#1}}
\makeatother

\newcommand{\gem}[1]{\mbox{\textsc{gem}}}

\newcommand{\hidetext}[1]{}
\newcommand{\ignore}[1]{}

\ifcomment
    \newcommand{\pinaforecomment}[3]{\colorbox{#1}{\parbox{.8\linewidth}{#2: #3}}}

    \newcommand{\prtodo}[1]{\pinaforecomment{lightblue}{pr}{#1}}
    \newcommand{\prtodoi}[1]{\pinaforecomment{lightblue}{pr}{#1}}
\else
    \newcommand{\pinaforecomment}[3]{}
    \newcommand{\prtodo}[1]{}
    \newcommand{\prtodoi}[1]{}
\fi

\newcommand{\smallurl}[1]{ \begin{tiny}\url{#1}\end{tiny}}

\definecolor{lightblue}{HTML}{3cc7ea}
\definecolor{CUgold}{HTML}{CFB87C}
\definecolor{grey}{rgb}{0.95,0.95,0.95}
\definecolor{ceil}{rgb}{0.57, 0.63, 0.81}
\definecolor{UMDred}{HTML}{ed1c24}
\definecolor{UMDyellow}{HTML}{ffc20e}

\usepackage{caption}
\usepackage{subcaption}
\usepackage{float}
\usepackage{soul}

\usepackage{times}
\usepackage{latexsym}

\usepackage[T1]{fontenc}

\usepackage[utf8]{inputenc}

\usepackage{microtype}

\usepackage{inconsolata}
\usepackage{enumitem}
\setitemize{noitemsep,topsep=0pt,parsep=0pt,partopsep=0pt}

\title{Large Language Models Help Humans Verify Truthfulness--- \\ Except When They Are Convincingly Wrong}

\author{Chenglei Si$^{1}$ \hspace{0.3cm} Navita Goyal$^{2}$ \hspace{0.3cm} Sherry Tongshuang Wu$^{3}$   \\
 \textbf{Chen Zhao}$^{4}$ \hspace{0.1cm}  \textbf{Shi Feng}$^{5}$ \hspace{0.1cm}  \textbf{Hal Daumé III}$^{2,6}$ \hspace{0.1cm}   \textbf{Jordan Boyd-Graber}$^{2}$\\
  $^{1}$Stanford University \hspace{0.4cm}
  $^{2}$University of Maryland \hspace{0.4cm}
  $^{3}$Carnegie Mellon University
  \\
  $^{4}$NYU Shanghai \hspace{0.6cm} 
  $^{5}$New York University \hspace{0.6cm}
  $^{6}$ Microsoft Research \\
  \texttt{clsi@stanford.edu} \\
}

\begin{document}
\maketitle
\begin{abstract}
  Large Language Models (LLMs) are increasingly used for accessing information on the web.
  Their truthfulness and factuality are thus of great interest. To help users make the right decisions about the information they get, LLMs should not only provide information but also help users fact-check it.
  Our experiments with 80 crowdworkers compare language models with search engines (information retrieval systems) at facilitating fact-checking.
  We prompt LLMs to validate a given claim and provide corresponding
  explanations.
  Users reading LLM explanations are significantly more efficient than those using search engines while achieving similar accuracy.
  However, they over-rely on the LLMs when the explanation is wrong.
  To reduce over-reliance on LLMs, we ask LLMs to provide contrastive information---explain both why the claim is true and false, and then we present both sides of the explanation to users.
  This contrastive explanation mitigates users’ over-reliance on LLMs, but cannot significantly outperform search engines.
Further, showing both search engine results and LLM explanations offers no complementary benefits compared to search engines alone.
Taken together, our study highlights that natural language explanations by LLMs may not be a reliable replacement for reading the retrieved passages, especially in high-stakes settings where over-relying on wrong AI explanations could lead to critical consequences.
\end{abstract}


\begin{figure*}[t]
\centering
  \includegraphics[trim={2cm 4cm 2cm 1.5cm},width=0.8\textwidth]{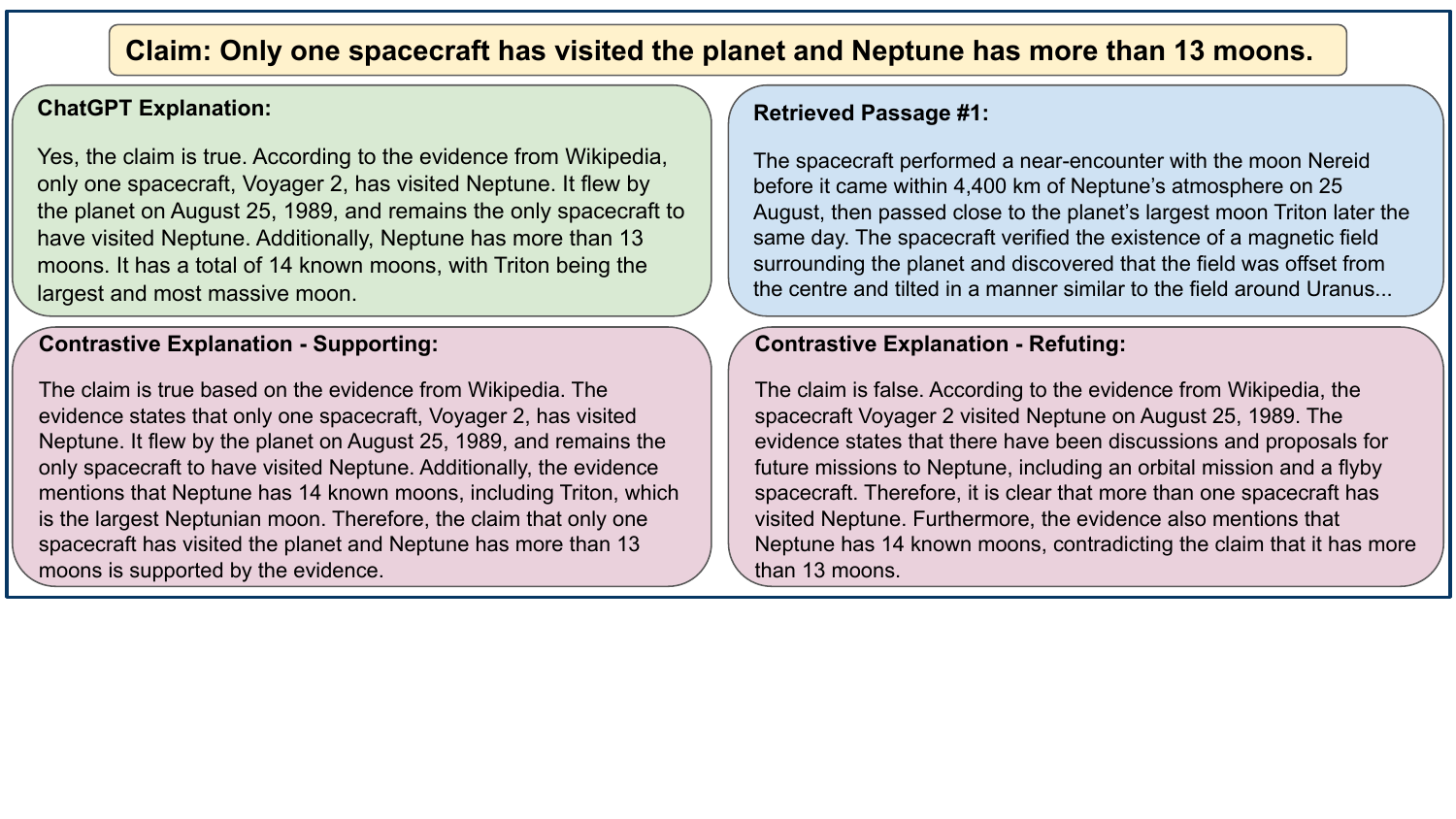}

  \caption{An example claim and the corresponding ChatGPT explanation, retrieved passages (abridged), and contrastive explanation. The claim is true and the refuting explanation has factual errors and reasoning contradiction.}
  
  \label{fig:example}
\end{figure*}

\section{Introduction}
Imagine you are told a claim about Neptune: \textit{``Only one
spacecraft has visited Neptune and it has more than 13 moons.''} and
you want to verify whether it is factual.
What would you do---look up
relevant pages from search engines or ask ChatGPT for its take?
%
This is not just a question of checking a piece of trivia; our
information ecosystem depends on people being able to check the
veracity of information online.
Misinformation, whether accidental or deliberate, has the potential to
sway public opinion, influence decisions, and erode trust in credible
sources~\cite{Faris2017PartisanshipPA,Mendes2017TroopsTA}.
Moreover, the wide adoption of large language models like ChatGPT
increases the danger of misinformation, both by malicious actors and
models generating inadvertent hallucinations~\cite{Pan2023OnTR}.


Consequently, verifying the accuracy of information is important.
Fact-checking claims is a well-established task in
NLP~\cite{Thorne2018FEVERAL,Guo2021ASO}.
However, automated fact-checkers are far from perfect, and they are
only useful when users trust their
predictions~\cite{Nakov2021AutomatedFF}.
Building that trust and providing effective help is crucial: a team
without trust leads to suboptimal human--AI team performance while
over-trusting wrong AI predictions could lead to catastrophic failures
in high-stakes applications.
Therefore, in real-life applications, we care about the AI-assisted
human accuracy of fact-checking, rather than evaluating and improving
automated fact-checkers alone~\cite{HCAI}.

The two major types of tools for helping human users (many of which are non-expert fact-checkers)
are \emph{retrieval} and \emph{explanation}~\cite{Nakov2021AutomatedFF}, exemplified by the widely-used web search engines (\textit{e.g.}, Google) and generative language models (\textit{e.g.}, ChatGPT) respectively. 
Showing retrieved passages to users has long been established as an effective information-seeking tool~\cite{Vlachos2014FactCT}. 
%
%
In contrast, the usefulness of generative explanations on fact-checking remains understudied.
On the one hand, competent generative models (especially LLMs) can generate fluent and convincing-looking natural language explanations that not only provide an answer (\textit{i.e.}, whether the claim is true or false), but also elucidate the context and basis of its judgment. 
On the other hand, these models are prone to hallucinations~\cite{Min2023FActScoreFA,Liu2023EvaluatingVI}, so the users are frequently left to their own devices.

This work studies \emph{whether language models can assist fact-checking}.
To contrast whether explanations or retrieved pages from a search engine are better, we compare LLMs with retrieval models mimicking a search engine experience and experiment with ways where retrieved passages can be paired with explanations, aiming to provide a practical guide on what is the most helpful tool. 
%
We base our evaluation on FoolMeTwice~\cite{Eisenschlos2021FoolMT},
 an adversarial dataset with interesting claims crowdsourced and gold evidence from Wikipedia~\cite{Eisenschlos2021FoolMT}. 
Participants verify whether the claim is factually true or false:
Figure~\ref{fig:example} shows an example to illustrate the
explanation and retrieved passages that participants see.

In our study, showing explanation and retrieved passages lead to similar human accuracy ($74\%$ and $73\%$ respectively) on difficult-to-verify claims ($59\%$ without AI assistant), but reading natural language explanations is significantly faster ($1.01$ min/claim vs $2.53$ min/claim). However, humans over-trust ChatGPT explanations where they agree with the explanation most of the time, even when the explanation is \emph{wrong}. 

To combat over-reliance on natural language explanations, we explore two improvements: 1) contrastive explanations---present both supporting and refuting arguments generated by ChatGPT to the user---and 2) combining retrieval and explanation (showing both to users). 
Both methods significantly reduce over-reliance on wrong AI explanations. 
However, their user fact-checking accuracy is no better than just showing users the retrieved passages. 
%
Overall, our work underscores the potential benefit and danger of natural language explanations as a tool in the battle against misinformation. They can save time, but at the same time the difficulty of combating over-reliance and the redundancy when combining retrieval and explanation remain.
Turning back to the question of what users should do to verify factuality: taking longer time to read the retrieved passages is still more reliable!

\section{Related Work}


We review relevant literature from NLP and HCI on fact-checking, explanations, and over-reliance. We also discuss additional related work in Section~\ref{sec:more_related_work}

\subsection{Fact Checking}

Abundant datasets have been collected for training and evaluating automatic fact-checking models, such as FEVER~\cite{Thorne2018FEVERAL,Schuster2021GetYV,Guo2021ASO} and SciFact~\cite{Wadden2020Scifact}. 
Various techniques improve the fact-checking pipeline, such as jointly reasoning across evidence articles and claims~\cite{Popat2018DeClarEDF}, and
breaking complex claims into atomic sub-claims~\cite{Min2023FActScoreFA,Kamoi2023WiCERE}.
Instead of improving automatic fact-checking \emph{per se}, we focus on how to improve human fact-checking via user studies. 

Compared to automated approaches, there are relatively few prior user studies. Notably, \citet{Fan2020GeneratingFC} synthesized summaries for retrieved passages to improve efficiency for users and \citet{Robbemond2022UnderstandingTR} compared showing explanations in different modalities to users. However, the advent of LLMs such as ChatGPT make it possible to generate plausible natural language explanations, and we are the first work to systematically evaluate such explanations in comparison to conventional retrieval methods.

\subsection{Explainable AI}

%
A thread of work in explainable AI (XAI) attempts to generate useful explanations in various formats~\cite{Wiegreffe2021TeachMT}, such as highlighting~\cite{Schuff2022HumanIO}, 
 feature importance~\cite{Ribeiro2016Lime}, free-text rationales~\cite{Ehsan2018Rationale}, and structured explanations~\cite{Lamm2020QEDAF}. 
%
As one end goal of explanations is to aid human verification of AI predictions and inform decision-making~\cite{Vasconcelos2022ExplanationsCR,Fok2023Verifiability}, much work in XAI literature has focused on human-centered evaluation of explanations~\cite{Sangdeh2021Manipulating}.
%
%
Closest to our work, \citet{Feng2018WhatCA} evaluated human-AI collaborative Quizbowl question answering and compared the effectiveness of showing retrieved passages, highlighting, and showing multiple guesses made by the system. 
This previous work used only a retrieval component, while our new approach allows us to directly compare ChatGPT-generated explanations (in the form of free-text rationales) with retrieved passages for aiding claim verification and explore whether natural language explanations and retrieved evidence yield complementary benefits. 
\citet{Joshi2023AreMR} studied free-text explanations in question-answering setting: their rationales do not help users much, especially when the rationales are misleading. In contrast to their work, we \emph{contrast} model-generated explanations with passages retrieved from external sources (Wikipedia). 

\subsection{Trust Calibration and Over-Reliance}

Existing work has identified the issue of human over-reliance on AI predictions across various application settings, where humans tend to trust AI predictions \emph{even when they are wrong}~\cite{Bussone2015Role,Lai2021TowardsAS}. 
A growing line of work attempts to mitigate such over-reliance, for example by providing explanations~\cite{Bansal2020DoesTW,Zhang2020EffectOC,Mohseni2020MachineLE,Vasconcelos2022ExplanationsCR,Das2022ProtoTExEM}, communicating model uncertainty~\cite{Prabhudesai2023UnderstandingUH,Si2022ReExaminingCT}, showing AI model accuracy~\cite{Yin2019UnderstandingTE}, and prompting slow thinking~\cite{Buccinca2021ToTrust} to help users calibrate their trust. 
Our work also contributes to this line of work by 
revealing over-reliance in fact-checking. We propose new ways of potentially combatting over-reliance including contrastive explanation and combining explanation with retrieval.

\section{Research Questions}

To understand the comparative advantages of retrieval and explanation in human fact verification, we pose the following research questions:

\begin{itemize*}
    \item \textbf{RQ1}: Are natural language explanations more effective than retrieved passages for human fact-checking?

    \item \textbf{RQ2}: Can contrastive explanations---arguing for or against a fact being true---mitigate over-reliance and be more effective than non-contrastive explanations? 

    \item \textbf{RQ3}: Are there complementary benefits in presenting both natural language explanations and retrieved passages? 
\end{itemize*}

%
We investigate these questions through a series of human studies: we show participants claims that need to be verified, potentially aiding them with different pieces of evidence (\autoref{fig:example}). This is a between-subjects study; thus, we vary the evidence presented to participants in different conditions:
%
\begin{itemize*}
    \item \textbf{Baseline}: We show users only the claims without any additional evidence.

    \item \textbf{Retrieval}: We show the top 10 paragraphs retrieved from Wikipedia along with the claim to be verified.

    \item  \textbf{Explanation}: We show the ChatGPT~\footnote{We use \texttt{gpt-3.5-turbo} in all experiments.} explanation along with the claim. 

    \item \textbf{Contrastive Explanation}: We present users ChatGPT's supporting and refuting arguments side by side.

    \item \textbf{Retrieval + Explanation}: We present both the retrieved passages as well as the (non-contrastive) natural language explanations to users. 
\end{itemize*}
%
%
In the \textbf{Explanation} and \textbf{Retrieval + Explanation} conditions, the ChatGPT prediction on whether the claim is true or false is part of the explanation,
while in the other conditions, users only see the evidence but not the prediction. 

%

\section{Study Design Overview}

We present an overview of the study design including the task setup, data used, measured variables, models used, and users involved. 

\subsection{Task, Data, and Variables}
\label{sec:dataset}


We ask human annotators to look at claims and decide whether it is true or false. 
We use the FoolMeTwice dataset~\cite{Eisenschlos2021FoolMT} over other claim-verification datasets because FoolMeTwice is adversarial: crowdworkers write claims based on Wikipedia to maximally fool another set of annotators whose task is to verify these claims. 
This ensures that all the claims are hard to verify, mimicking potential real-world fake news arms race. 
%
For our human studies, we create a test set by randomly sampling 200 claims where half are true and half are false. To ensure that the selected claims are sufficiently complex,  we only sample claims requiring at least two different sentences from Wikipedia to verify. 



We sample 20 claims (half true and half false) for each participant to verify and randomize their order. 
For each claim, we ask for the participant's binary decision of whether they think the claim is true or false. We measure the accuracy of human decisions given that we know the gold labels of these claims. 
We also ask for the participant's confidence in their judgment on a scale of 1 to 5, and record the time used for verifying each claim.
We also ask for a free-form response of how the annotator makes their judgments. Appendix~\ref{sec:interface} and Figure~\ref{fig:interface} illustrate the interface setup.

\subsection{Retriever}



For the \textbf{Retrieval} and \textbf{Retrieval + Explanation} conditions, we show users the most relevant passages from Wikipedia. 
Specifically, we adopt a similar retrieval setup as \citet{Min2023FActScoreFA}, where we use the state-of-the-art Generalizable T5-based Retriever (GTR-XXL), an unsupervised dense passage retriever~\cite{Ni2021LargeDE}. 
We retrieve the top $10$ most relevant paragraphs from Wikipedia, where each paragraph has an average length of $188$ words. 
To measure the retrieval quality, we report two metrics on our test set. The full recall measures how often the top $10$ retrieved passages contain all evidence sentences required to verify the claim, which is $81.5\%$; partial recall measures how often the top 10 retrieved passages contain at least one evidence sentence required to verify the claim, which is $93.0\%$. 

\subsection{Explanation Generation}


We study two types of natural language explanations with ChatGPT: non-contrastive explanation and contrastive explanation. 
In the \textbf{Explanation} and \textbf{Retrieval + Explanation} conditions, we generate \textbf{non-contrastive} explanations, where we construct the prompt by concatenating the top $10$ retrieved passages, followed by the claim to be verified, then appending the question \textit{``Based on the evidence from Wikipedia, is the claim true? Explain in a short paragraph.''} We measure the accuracy of these explanations by manually extracting the answer (true or false) from the explanations and comparing with the gold labels. ChatGPT-generated explanations achieve an accuracy of $78.0\%$ (judged based on the AI predictions only, not the reasoning processes). 
In the \textbf{Contrastive Explanation} condition, 
we prompt ChatGPT to generate both a supporting answer and a refuting answer. Specifically, after concatenating the retrieved passages and the claim, we append two different questions: 1) \textit{``Based on the evidence from Wikipedia, explain in a short paragraph why the claim is \textbf{true}.''} and 2) \textit{``Based on the evidence from Wikipedia, explain in a short paragraph why the claim is \textbf{false}.''} We then show both of these generated explanations to annotators, which functions similarly to a single-turn debate~\cite{Parrish2022SingleTurnDD,Michael2023DebateHS}. 

Additionally, in \textbf{Retrieval + Explanation}, we automatically insert citations to the explanation text to attribute the arguments to corresponding retrieved passages. 
This is through prompting ChatGPT with a manually crafted example of inserting citations into the explanations based on the retrieved passages, which enables citations in language model generation~\cite{Gao2023EnablingLL}.
%
For all cases, we ground the explanation generation on the retrieved passages. This is because grounding significantly improves the accuracy of explanations. For example, for non-contrastive explanations, grounding improves the accuracy from $59.5\%$ to $78.0\%$. For all cases, we use a temperature value of $0$ for ChatGPT generation to minimize randomness. 

\subsection{Users}

We recruit participants from Prolific.
We recruit $16$ annotators for each condition and each annotator verifies $20$ claims, resulting in $20 \times 16 \times 5 = 1500$ annotations across all five conditions. We compensate all annotators at least $\$14$ per hour.
Our study is approved by the University of Maryland Institutional Review Board. 


\section{Experiment Results}

Next, we present and discuss experiment results addressing each of the three research questions, as well as the empirical findings on over-reliance.

\begin{figure}[t]
     \begin{subfigure}[b]{0.235\textwidth}
         \centering
         \includegraphics[trim={0.5cm 0.5cm 0.5cm 0.5cm},width=\textwidth]{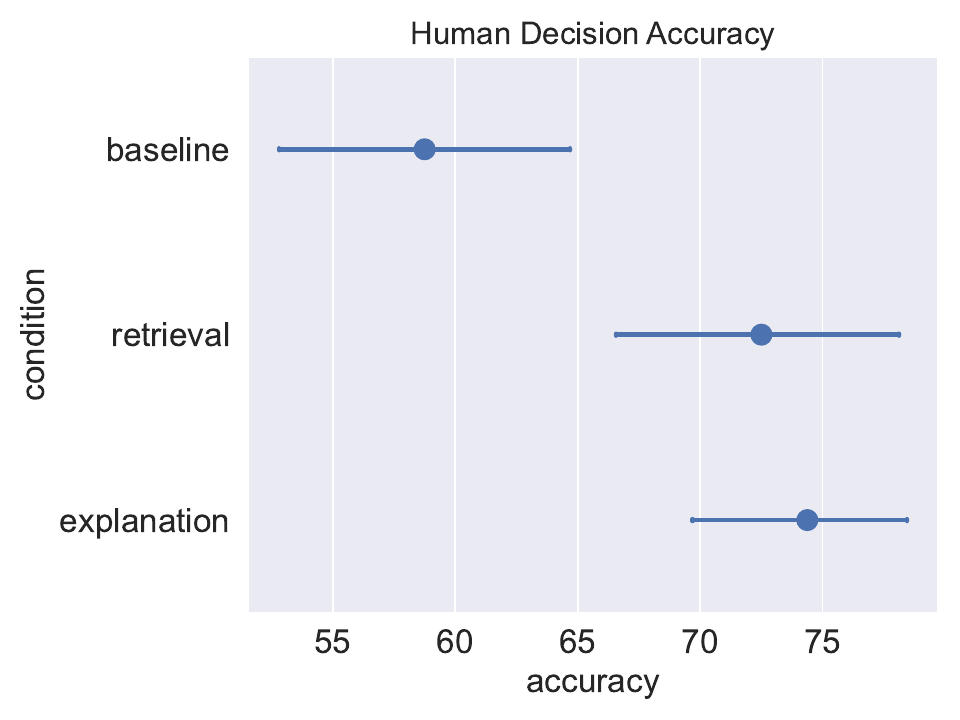}
         \caption{Accuracy}
         \label{fig:experiment_1_accuracy}
     \end{subfigure}
     \hfill
     \begin{subfigure}[b]{0.235\textwidth}
         \centering
         \includegraphics[trim={0.5cm 0.5cm 0.5cm 0.5cm},width=\textwidth]{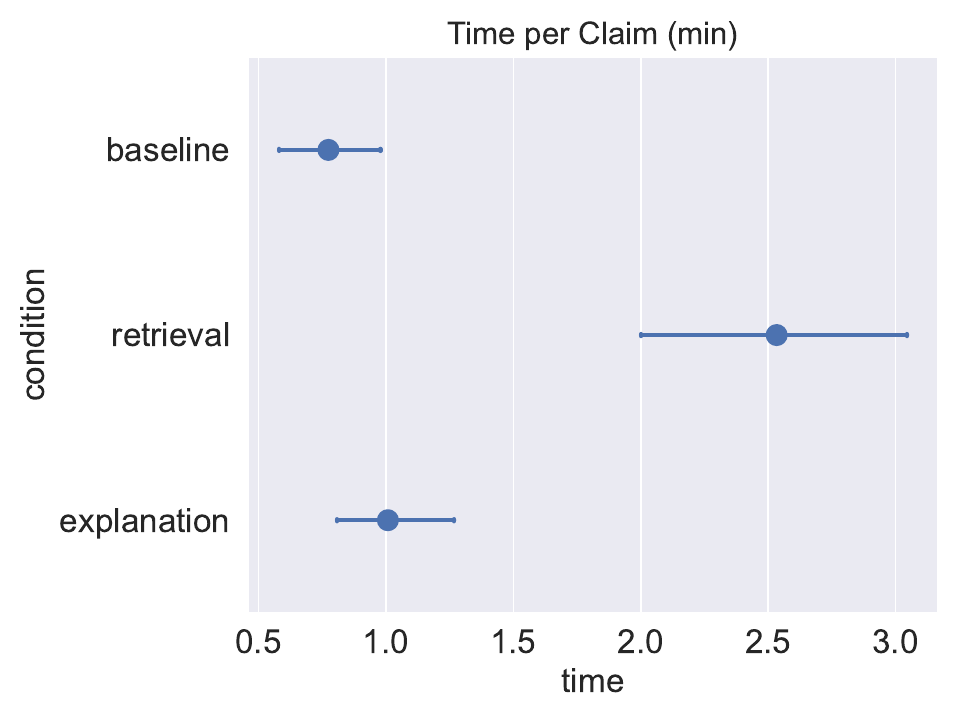}
         \caption{Time}
         \label{fig:experiment_1_time}
     \end{subfigure}
    \caption{
    Human decision accuracy and average time spent on verifying a claim. Both retrieval and explanation significantly improve human verification accuracy, while explanation takes a significantly shorter time.  
}    
\label{fig:experiment_1}
\end{figure}

\begin{figure}[t]
    \centering
     \begin{subfigure}[b]{0.235\textwidth}
         \centering
         \includegraphics[trim={0.5cm 0.5cm 0.5cm 0.5cm},width=\textwidth]{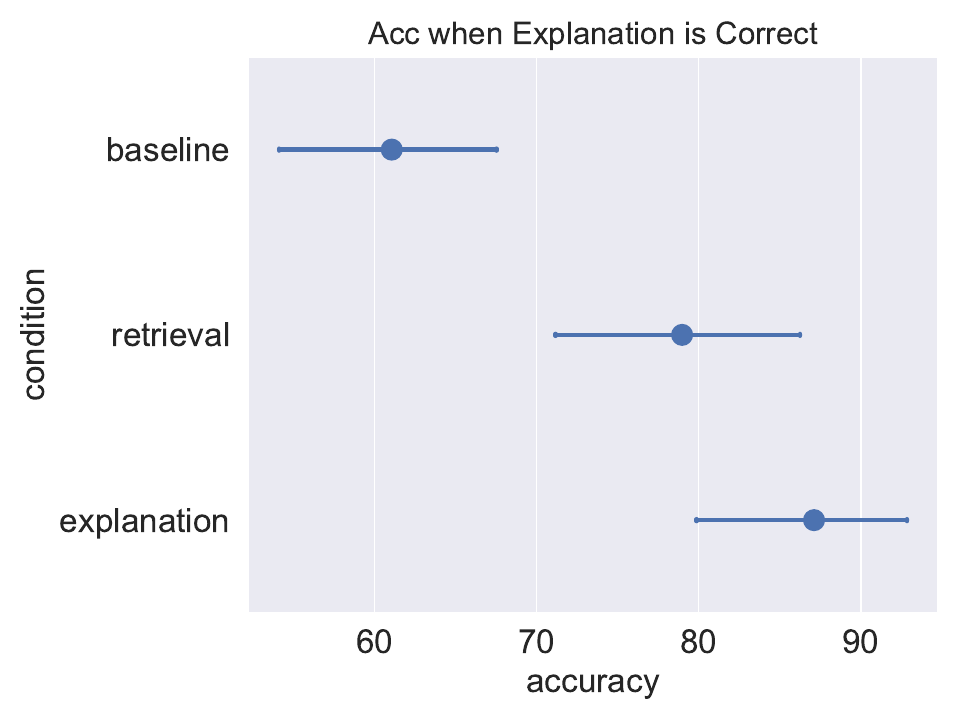}
         \caption{Human decision accuracy on examples where the explanation is correct.}
         \label{fig:analysis_1_correct}
     \end{subfigure}
     \hfill
     \begin{subfigure}[b]{0.235\textwidth}
         \centering
         \includegraphics[trim={0.5cm 0.5cm 0.5cm 0.5cm},width=\textwidth]{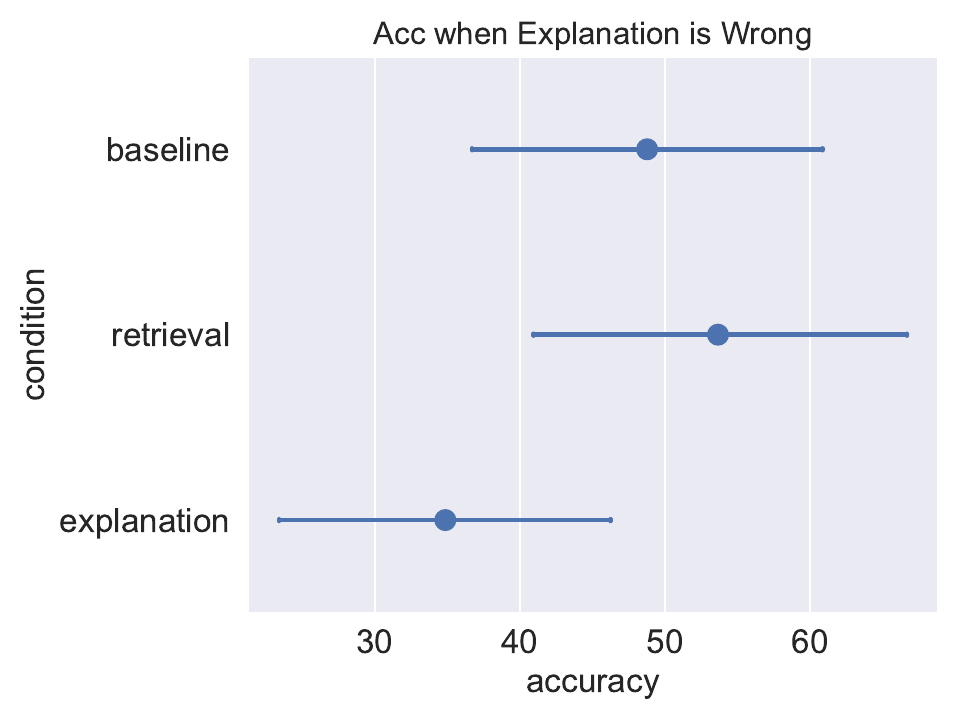}
         \caption{Human decision accuracy on examples where the explanation is wrong.}
         \label{fig:analysis_1_wrong}
     \end{subfigure}
    \caption{
    Human verification accuracy broken down into two subsets: examples on which the explanation gives the correct labels, and examples on which the explanation gives the wrong labels. Humans over-rely on explanations so that they achieve significantly lower accuracy than the baseline when the explanation is wrong. 
}    
\label{fig:analysis_1}
\end{figure}


\subsection{RQ1: Are natural language explanations more effective than retrieved passages for human fact checking?}







We compare three conditions: the \texttt{Baseline} condition (showing users only the claims); the \texttt{Retrieval} condition (showing the top 10 paragraphs retrieved from Wikipedia); and the \texttt{Explanation} condition (showing the ChatGPT explanation along with the claim). We do not set a time limit but record the time taken for each claim. 




\autoref{fig:experiment_1_accuracy} shows the AI-assisted human verification accuracy across conditions. We test the significance of our results using Student’s t-tests with Bonferroni correction~\footnote{We inspected all data with histograms and $Q-Q$ plots to verify that the data approximate normality before applying $t$-tests.}. We start with examining whether ChatGPT explanations and retrieved passages are indeed helpful for humans. 

\textbf{Showing ChatGPT explanation improves human accuracy.} 
When showing explanations to users, the accuracy is $\mu=0.74 \pm \sigma=0.09$ compared to the baseline condition where claims are shown without any additional evidence $(0.59 \pm 0.12)$. The improvement in accuracy is significant $(z=-4.08, p=0.00015)$.~\footnote{We use $p < 0.05$ as the threshold for all significance tests.}

\textbf{Showing retrieved passages improves human accuracy.}
When showing retrieved passages to users, they achieve the accuracy of $(0.73 \pm 0.12)$ as compared to the baseline condition where claims are shown without any additional evidence $(0.59 \pm 0.12)$. The improvement in accuracy is significant $(z=-3.15, p=0.0018)$. 
Now that both ChatGPT explanation and retrieved passages help humans more accurately verify claims, we examine their comparative advantages in both accuracy and time. 


\textbf{Showing ChatGPT explanations does not significantly improve accuracy over showing retrieved passages.} Comparing the accuracy in the explanation condition $(0.74 \pm 0.09)$ and the retrieval condition $(0.73 \pm 0.12)$, the improvement in accuracy is not significant $(z=-0.48, p=0.32)$.

However, \textbf{reading ChatGPT explanation is significantly faster than reading retrieved passages.} 
We compare the time taken to verify claims in \autoref{fig:experiment_1_time}. When verifying with retrieved passages, the time taken to verify each claim is $(2.53 \pm 1.07)$ minutes while for the explanation condition, it takes $(1.01 \pm 0.45)$ minutes. Showing explanations allows significantly faster decision time than showing retrieved passages $(z=-5.09, p=9.1e-6)$. 


\subsection{Breakdown Analysis: The Danger of Over-Reliance}

While ChatGPT explanations show promise in aiding human fact verification, 
the aggregate results obscure the danger when the explanation gives wrong answers. To examine what happens in those cases, we break down the analysis, manually annotating the ChatGPT explanation for each claim based on whether it gives the correct answer (whether the claim is true or false). We then split all user responses into two subsets: ones with correct answers from ChatGPT and ones where the ChatGPT explanation is wrong (\autoref{fig:analysis_1_correct} and~\autoref{fig:analysis_1_wrong}, respectively). 


\textbf{Users achieve the highest accuracy when the explanations are correct, but below-random accuracy when the explanations are wrong.}
When the explanation is correct, users' accuracy is $(0.87 \pm 0.13)$, higher than the baseline of having no evidence $(0.61 \pm 0.13)$ as well as the retrieval condition $(0.79 \pm 0.15)$. However, when the explanation is wrong, users tend to over-trust the explanations and only achieve an accuracy of $(0.35 \pm 0.22)$ as compared to the baseline condition $(0.49 \pm 0.24)$ and the retrieval condition $(0.54 \pm 0.26)$. 
Moreover, \textbf{users spend similar time on claims with correct and wrong explanations}, further indicating that they are not deliberately differentiating correct and wrong explanations and instead tend to trust most of the explanations. 
We also look at the free-form responses from users for their decision rationales, the most common responses include: (1) ChatGPT's explanation looks convincing, especially with quotes from the retrieved passages (even when the quotes or reasoning are wrong); (2) They do not have any prior knowledge on the topic so would just trust ChatGPT. 

\textbf{In comparison, retrieved passages suffer less from over-reliance.}
On examples where the ChatGPT explanations are correct, the retrieval condition achieves the accuracy of $(0.79 \pm 0.15)$, surpassing the baseline condition $(0.61 \pm 0.13)$. On examples where the ChatGPT explanations are wrong, the retrieval condition achieves the accuracy of $(0.54 \pm 0.26)$ compared to the baseline $(0.49 \pm 0.24)$. While there is still an accuracy drop in these examples, possibly because they are harder to verify, the performance discrepancy between the two cases (ChatGPT explanation correct vs wrong) is much less severe in the retrieval condition. 
This highlights the pitfall of using ChatGPT explanation to aid verification: users over-rely on the explanations, even when they are wrong and misleading. To combat this problem, we next explore two strategies for mitigation: contrastive explanation and combining retrieval and explanation.


\begin{figure}[t]
    \centering
     \begin{subfigure}[b]{0.235\textwidth}
         \centering
         \includegraphics[trim={0.5cm 0.5cm 0.5cm 0.5cm},width=\textwidth]{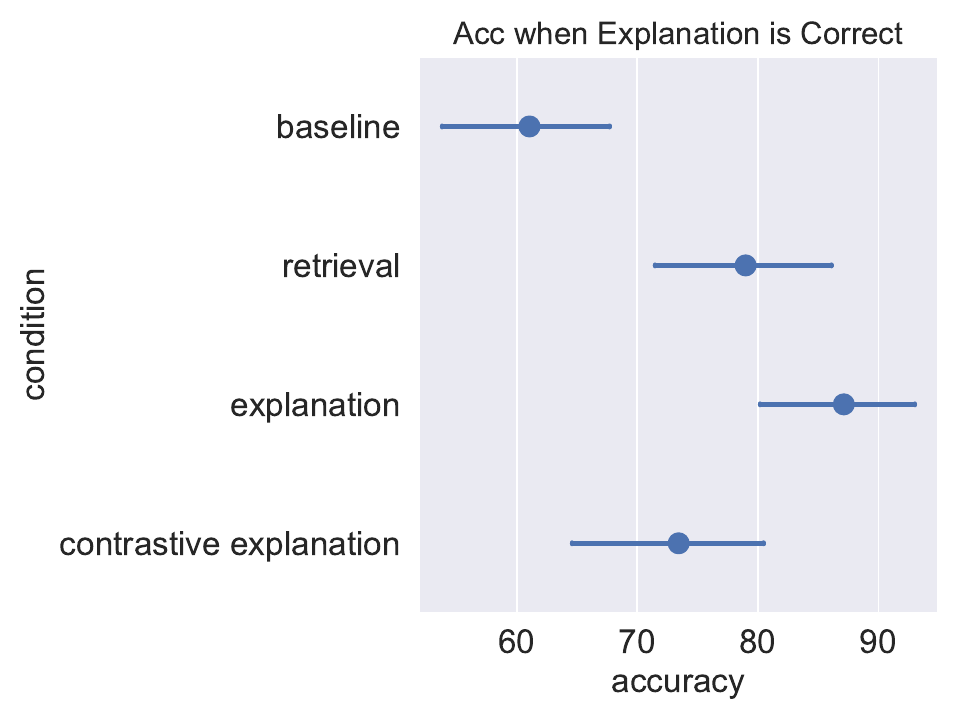}
         \caption{Human decision accuracy on examples where the explanation is correct.}
         \label{fig:experiment_2_accuracy_correct}
     \end{subfigure}
     \hfill
     \begin{subfigure}[b]{0.235\textwidth}
         \centering
         \includegraphics[trim={0.5cm 0.5cm 0.5cm 0.5cm},width=\textwidth]{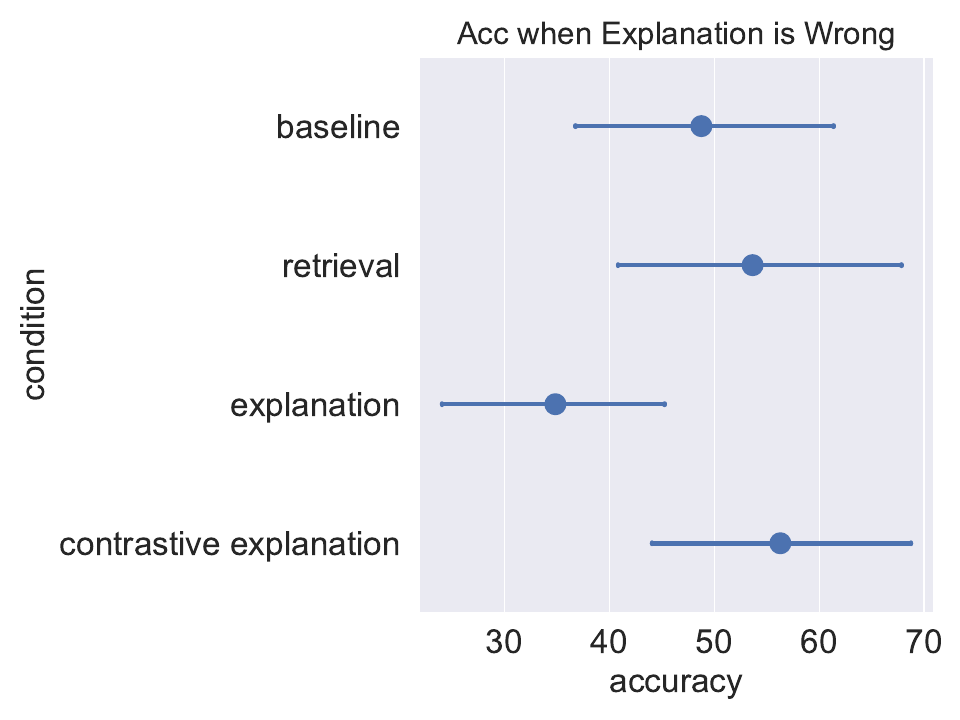}
         \caption{Human decision accuracy on examples where the explanation is wrong.}
         \label{fig:experiment_2_accuracy_wrong}
     \end{subfigure}
     \hfill
     \begin{subfigure}[b]{0.235\textwidth}
         \centering
         \includegraphics[trim={0.5cm 0.5cm 0.5cm 0cm},width=\textwidth]{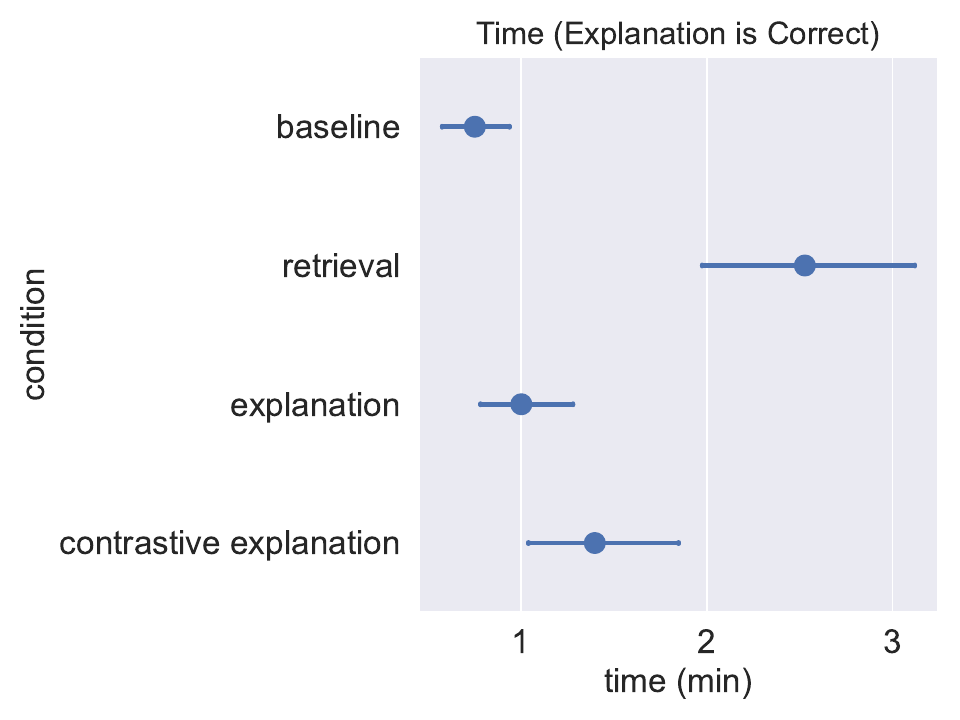}
         \caption{Human decision time on examples where the explanation is correct.}
    \label{fig:experiment_2_time_correct}
     \end{subfigure}
         \hfill
     \begin{subfigure}[b]{0.235\textwidth}
         \centering
         \includegraphics[trim={0.5cm 0.5cm 0.5cm 0cm},width=\textwidth]{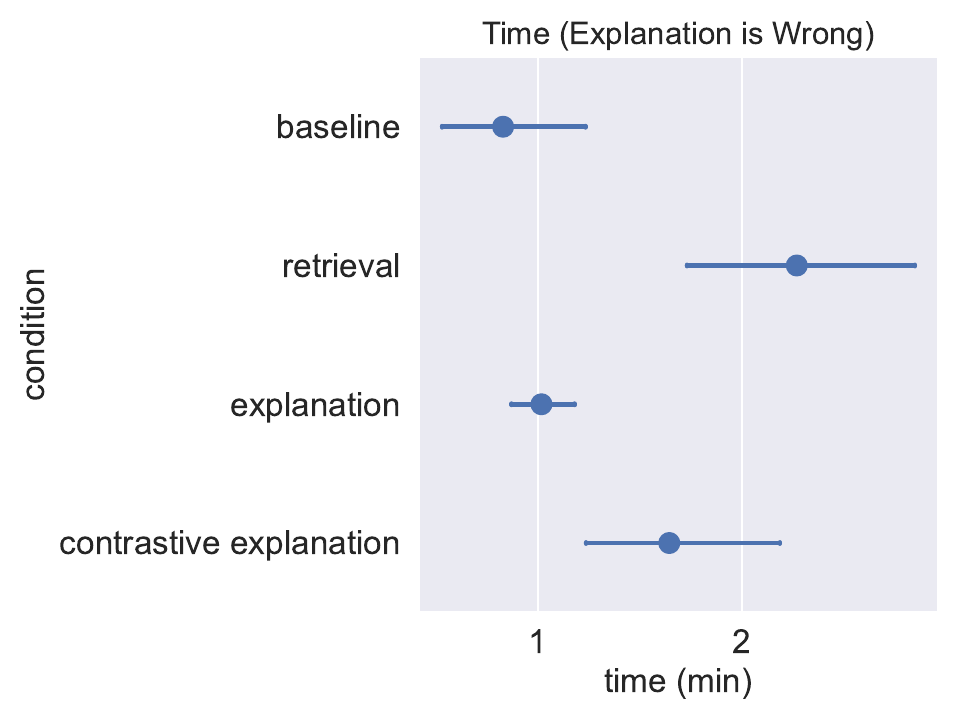}
         \caption{Human decision time on examples where the explanation is wrong.}
    \label{fig:experiment_2_time_wrong}
     \end{subfigure}
    \caption{
    Verification accuracy and time broken down by whether the (non-contrastive) explanation is correct. Contrastive explanation significantly improves accuracy over non-contrastive explanation on examples where the non-contrastive explanation is wrong, with some drop in accuracy on examples where the non-contrastive explanation is correct.  
}    
\label{fig:experiment_2}
\end{figure}

\subsection{RQ2: Can contrastive explanations mitigate over-reliance and be more effective than non-contrastive explanations?}







In addition to the three conditions from the previous section (\texttt{Baseline}, \texttt{Retrieval}, and \texttt{Explanation}), we additionally compare the \texttt{Contrastive Explanation} condition where we present users ChatGPT's supporting and refuting arguments side by side (\autoref{fig:experiment_2}).
%
%
We first compare contrastive explanation with non-contrastive explanation. 

\textbf{Contrastive explanation improves human accuracy more than non-contrastive explanation when the non-contrastive explanation is wrong.} When the non-contrastive explanation is wrong, humans accuracy is $(0.35 \pm 0.22)$ due to over-reliance, but when switching to contrastive explanation improves the accuracy to $(0.56 \pm 0.24)$, which is significantly higher $(z=-2.52, p=0.009)$. 
When analyzing the free-response rationales of human judgment, the most common patterns of how people make correct judgments based on contrastive explanations are: (1) The correct side of the explanation is more compelling or thorough; (2) The wrong side of the explanation contains factual errors and wrong reasoning; (3) Both sides of the explanations give the same answer (even though ChatGPT was prompted to explain why the claim is true and false in the two sides of explanations). 

However, \textbf{contrastive explanation has lower human accuracy than non-contrastive explanation when the non-contrastive explanation is correct.} When the non-contrastive explanation is correct, humans' accuracy is $0.87 \pm 0.13$, higher than contrastive explanation $(0.73 \pm 0.15)$, indicating a significant drop $(z=-2.56, p=0.008)$. 
Unlike in non-contrastive explanations---where users can just take the AI prediction as the answer---they have to \emph{decide} which of the two contrastive sides of the explanation is correct.  
This is often tricky because LLMs can generate convincing explanations even for the wrong statements. 
For example, given the false claim \textit{``Joe Torre was the manager of the New York Yankees and guided the team to four World Series championships, and ranks third all-time in MLB history with 2,326 wins as a manager.''}, ChatGPT generates the supporting explanation {\color{blue}\textit{``Yes, the claim is true. According to the evidence from Wikipedia, Joe Torre was the manager of the New York Yankees from 1996 to 2007. He also ranks third all-time in MLB history with 2,326 wins as a manager.''}} and generates the refuting explanation {\color{brown}\textit{``The claim is false. According to the evidence from Wikipedia, Joe Torre was the manager of the New York Yankees and guided the team to six pennants and four World Series championships. He ranks fifth all-time in MLB history with 2,326 wins as a manager, not third.''}} 
Torre ranks \emph{fifth} all-time in MLB history with 2,326 wins as a manager but ChatGPT still generated a convincing-looking explanation for the wrong side by hallucinating he ranks third all-time rather than fifth. As a result, some users were misled. 
Overall, contrastive explanation shows promise in reducing over-reliance but incurs a trade-off in accuracy when the non-contrastive explanation is correct. Next, we also compare contrastive explanation with retrieval.


\textbf{Contrastive explanation does not significantly improve human accuracy over retrieval.} On examples where the non-contrastive explanation is correct, human accuracy with contrastive explanation is $0.73 \pm 0.15$, lower than the accuracy in the retrieval condition $(0.79 \pm 0.15)$. 
On examples where the non-contrastive explanation is wrong, contrastive explanation has comparable human accuracy of $0.56 \pm 0.24$ compared to retrieval $(0.54 \pm 0.26)$, and the difference is not significant $(z=0.29, p=0.61)$. 
Therefore, in both cases, contrastive explanations do not significantly improve human accuracy over retrieval, despite the evidence that contrastive explanations can mitigate over-reliance compared to non-contrastive explanations. 

Apart from the above quantitative results, we also manually analyze free-form responses of user's rationales to understand how users decide with contrastive explanations. 
\textbf{Users mostly base their judgment on the relative strength of the two sides of the explanations} (\textit{i.e.}, is the supporting or refuting explanation more convincing) ($41.8\%$). Example user rationales include: \textit{``The refutation seems more logically sound.''} and \textit{``The support explanation seems like it's trying too hard to make the claim true, but the refute puts it more plain and simple and makes more sense.''} 
\textbf{Sometimes both sides converge on the same answer} ($26.9\%$) and users would just adopt the consensus. For example, for the false claim \textit{``The only verified original sled prop from Citizen Kane was sold at a price of over a hundred thousand dollars.''}, users report \textit{``Both sides acknowledge that there were more than 1 sled prop, therefore refuting the claim.''}, even though the ChatGPT supporting explanation said \textit{``The claim is true.''} In several cases, ChatGPT would simply say the claim is true even though we prompt it for a refuting explanation (and vice versa), giving users a clear cue that the model could not make a strong argument for the wrong side.


\begin{figure}[t]
    \centering
     \begin{subfigure}[b]{0.235\textwidth}
         \centering
         \includegraphics[trim={0.5cm 0.5cm 0.5cm 0.5cm},width=\textwidth]{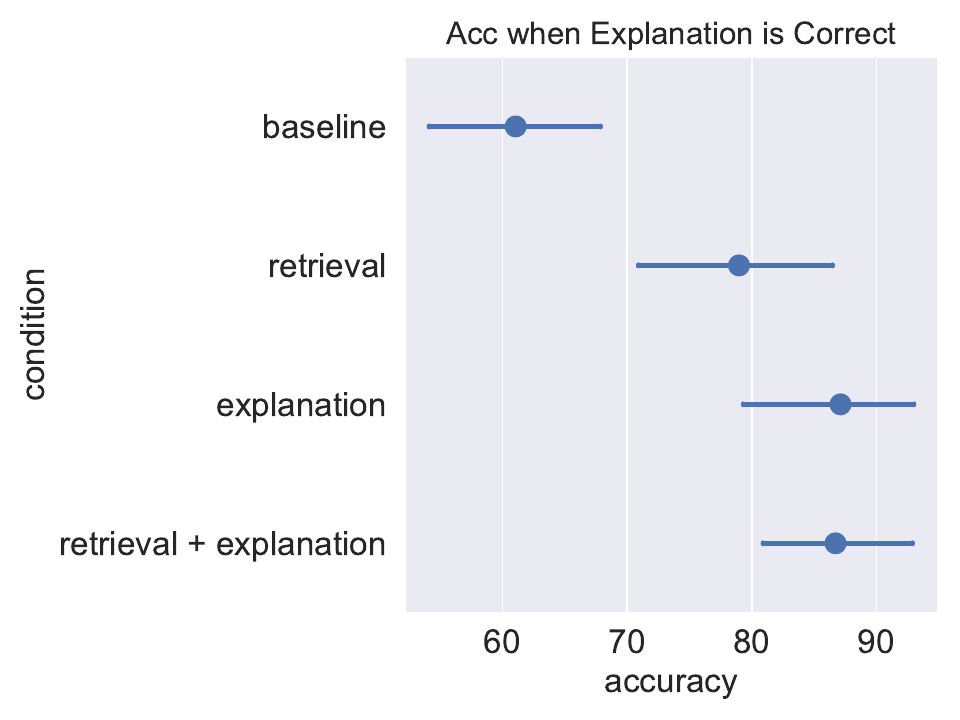}
         \caption{Human decision accuracy on examples where the explanation is correct.}
         \label{fig:experiment_3_accuracy_correct}
     \end{subfigure}
     \hfill
     \begin{subfigure}[b]{0.235\textwidth}
         \centering
         \includegraphics[trim={0.5cm 0.5cm 0.5cm 0.5cm},width=\textwidth]{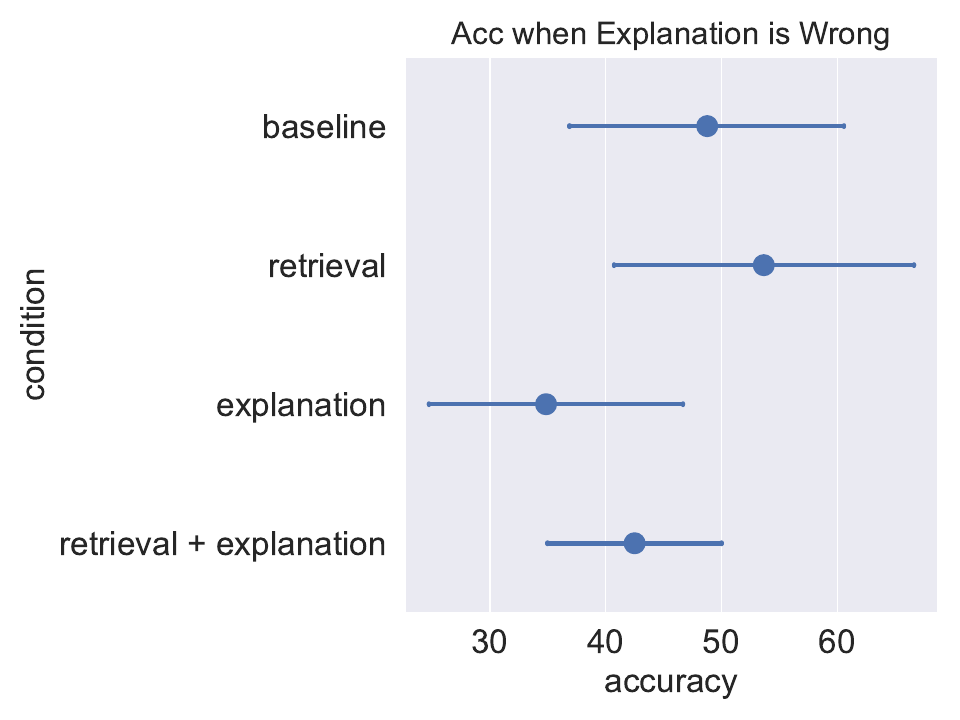}
         \caption{Human decision accuracy on examples where the explanation is wrong.}
    
         \label{fig:experiment_3_accuracy_wrong}
     \end{subfigure}
     \hfill
     \begin{subfigure}[b]{0.235\textwidth}
         \centering
         \includegraphics[trim={0.5cm 0.5cm 0.5cm 0cm},width=\textwidth]{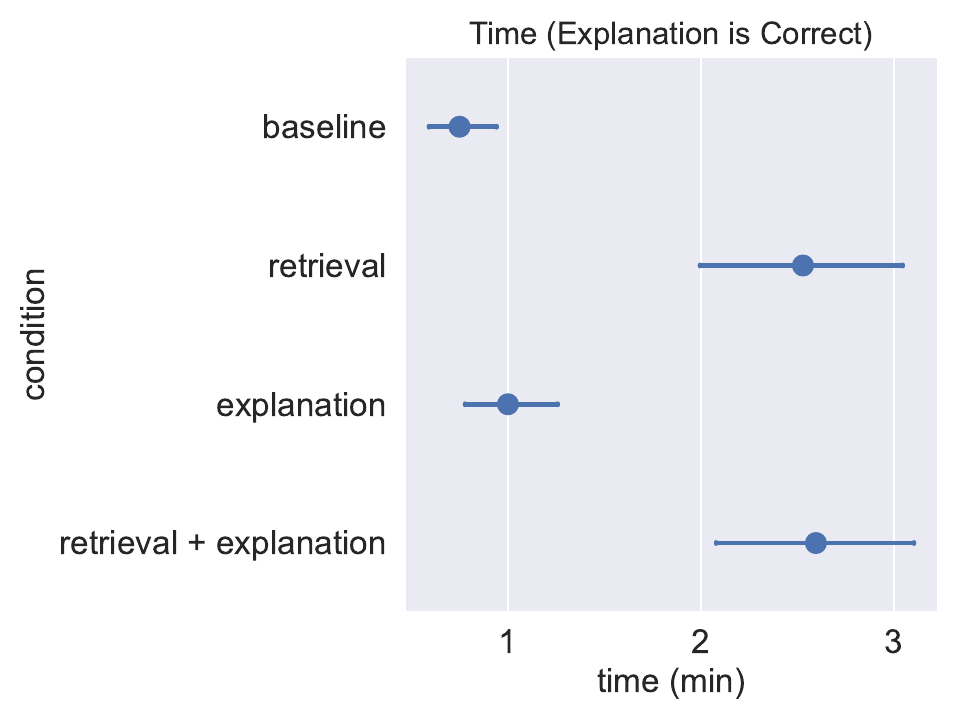}
         \caption{Human decision time on examples where the explanation is correct.}
    \label{fig:experiment_3_time_correct}
     \end{subfigure}
         \hfill
     \begin{subfigure}[b]{0.235\textwidth}
         \centering
         \includegraphics[trim={0.5cm 0.5cm 0.5cm 0cm},width=\textwidth]{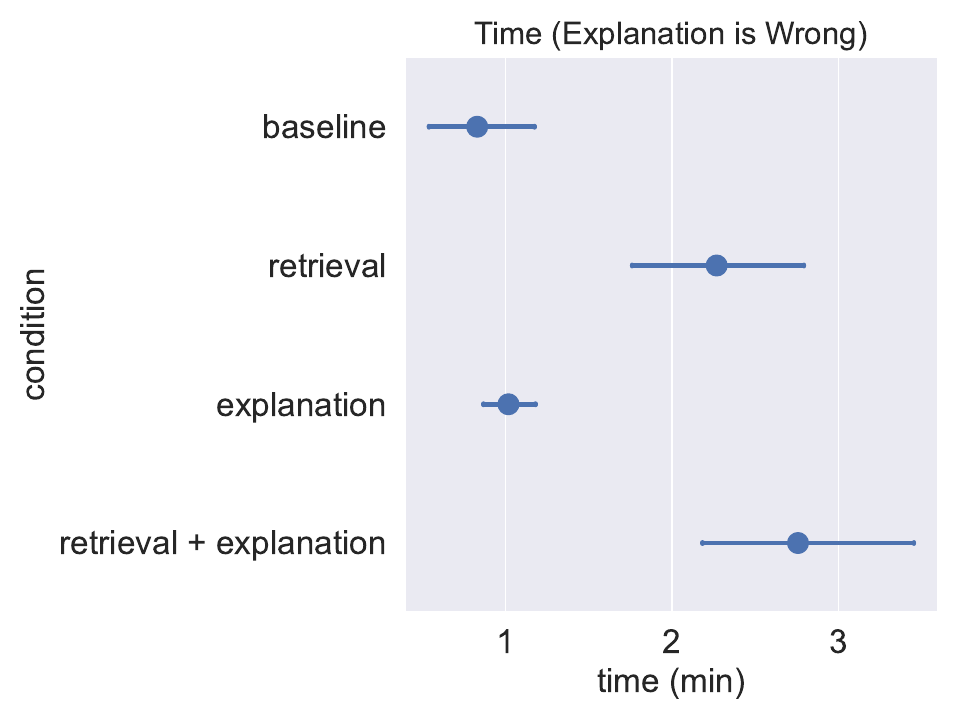}
         \caption{Human decision time on examples where the explanation is wrong.}
    \label{fig:experiment_3_time_wrong}
     \end{subfigure}
    \caption{
    Verification accuracy and time breakdown. Combining retrieval and explanation is not significantly better than just showing retrieved passages alone. 
}    
\label{fig:experiment_3}
\end{figure}

\subsection{RQ3: Are there complementary benefits in presenting both natural language explanations and retrieved passages? }






Apart from the \texttt{Baseline}, \texttt{Retrieval}, and \texttt{Explanation} conditions from earlier, we also compare with the (\texttt{Retrieval + Explanation}) condition where we present both to users (\autoref{fig:experiment_3}). 
%
%
We start by comparing whether combining explanation with retrieval is better than explanation alone. 

\textbf{Combining retrieval and explanation does not significantly improve accuracy over \texttt{explanation} when the explanation is correct.} 
When the explanation is correct, users' accuracy $(0.87 \pm 0.13)$ with explanations alone is comparable---i.e., no significant difference $(z=0.084, p=0.53)$---to retrieval combined with explanation $(0.87 \pm 0.12)$.

\textbf{Combining retrieval and explanation does not significantly improve accuracy over \texttt{explanation} alone when the explanation is wrong either.} When the explanation is wrong, users' accuracy $(0.35 \pm 0.22)$ in the explanation condition is slightly lower than combining retrieval and explanation $(0.43 \pm 0.16)$. The advantage of combining retrieval and explanation is not significant $(z=-1.06, p=0.15)$. 
Taken together, combining explanation and retrieval is not better than explanation alone. Next, we compare whether combining explanation with retrieval is better than retrieval alone. 

\textbf{Combining retrieval and explanation does not significantly improve accuracy over \texttt{retrieval} in cases where the explanation is correct.}
When the explanation is correct, users' accuracy in the \texttt{retrieval} condition $(0.79 \pm 0.15)$ is lower than combining both retrieval and explanation $(0.87 \pm 0.12)$.
There is a slight advantage in combining retrieval and explanation in this setting but the advantage is not significant $(z=-1.48, p=0.07)$.

\textbf{Combining retrieval and explanation does not significantly improve accuracy over \texttt{retrieval} in cases where the explanation is wrong.} 
When the explanation is wrong, users' accuracy $(0.54 \pm 0.26)$ in the \texttt{retrieval} alone condition beats combining both retrieval and explanation $(0.43 \pm 0.16)$, indicating a drop in accuracy in this case when combining retrieval and explanation. This means that combining retrieval and explanation offers no complementary benefits compared to retrieval alone. To understand whether users indeed read both the explanation and retrieved passages, we further compare their reading time. 



\textbf{Combining retrieval and explanation takes a longer time.}  In the \texttt{retrieval} condition, users take $2.5 \pm 1.1$ minutes to verify a claim; in the explanation condition, users take $1.0 \pm 0.4$ minutes to verify a claim; in the \texttt{retrieval + explanation} condition, users take $2.7 \pm 1.0$ minutes to verify a claim, indicating that combining retrieval and explanation increases the verification time, so users indeed spend time reading the explanation and retrieved passages. 
Moreover, in analyzing the free-form responses, the majority of the users base their judgment on the retrieved passages since the ChatGPT explanations are not always credible, further indicating that presenting ChatGPT explanations grounded on the retrieved passages does not really offer additional benefits than just presenting the retrieved passages themselves. 
Overall, combining retrieval and explanation might be redundant and inefficient.


\section{Meta-Analysis}

We analyze several additional factors across all experiments, such as the impact of retrieval recall, time, confidence, and qualitative responses. 


\subsection{Impact of Retrieval Recall}
\label{sec:retrieval}


We split examples into two categories: the first group where the top-10 retrieved passages contain all the necessary evidence to verify the claim (\textit{i.e.,} full recall $r=1$), and the second group where not all evidence is retrieved within the top-10 passages (\textit{i.e.,} full recall $r=0$). We analyze how the retrieval recall affects both the explanation accuracy as well as the human decision accuracy. 

\textbf{The explanation accuracy is much lower when the retrieval recall is low.} Over the entire test set of 200 examples, with full recall $r=1$, the explanation accuracy is $80.4\%$; when the full recall $r=0$, the explanation accuracy is $67.6\%$. This indicates that retrieval quality has a high impact on explanation accuracy, which in turn affects human decision accuracy. 

\textbf{Human decision accuracy is much lower when the retrieval recall is low.} 
In all cases (apart from the \texttt{Baseline} condition where users do not see any evidence), the human decision accuracy is lower when the full retrieval recall is 0, sometimes it is lower than the case of full recall $r=1$ by large margins, \textit{e.g.}, in the \texttt{Retrieval} condition and the \texttt{Retrieval + Explanation} condition (\autoref{fig:analysis_recall}).


\subsection{Correlation Between Accuracy and Time}
\label{sec:correlation}

There is little correlation between accuracy and time ($r = 0.099$, \autoref{fig:acc_time_corr}). 
Broken down for each condition, the correlations remain weak:
\begin{itemize*}
    \item Baseline: $r = -0.180$;
    \item Retrieval: $r = 0.089$;
    \item Explanation: $r = -0.539$;
    \item Contrastive Explanation: $r = -0.008$; and
    \item Retrieval + Explanation: $r = 0.148$.
\end{itemize*}

\subsection{Confidence Calibration}
\label{sec:confidence}

We convert users' confidence levels into discrete values $\mathcal{C} = \{0, 0.25, 0.5, 0.75, 1.0\}$.
Our goal is for users to have high confidence in their correct judgments and low confidence in their wrong judgments. We plot their average confidence on correct and wrong judgments in \autoref{fig:experiment_2_confidence}. 
User confidence is always low in the \texttt{Baseline} condition, which is reasonable since they do not have additional supporting evidence and are mostly making educated guesses. 
On correct judgments, users generally have high confidence (above $0.6$). However, \textbf{users are over-confident on wrong judgments}, with average confidence above $0.6$. 
The \texttt{Explanation} and \texttt{Contrastive Explanation} conditions incur lower user confidence on both correct and wrong judgments compared to the \texttt{Retrieval} condition, as well as the \texttt{Retrieval + Explanation} condition. 
These results highlight the difficulty of appropriately calibrating users' judgments.

\subsection{When Do Users Disagree with Explanations} 
\label{sec:qual_analysis}

We manually analyze the free-form decision rationales provided by users to
understand when they would disagree with ChatGPT.

\paragraph{(1) \textbf{How do users make the correct decision when the explanations are wrong?} }
\begin{itemize*}
    \item In \texttt{Explanation}, most users rely on \textbf{self-contradiction} ($40.7\%$). For example, given the true claim \textit{``Charles Evans Hughes shuffled off this mortal coil in Massachusetts, and then was taken to New York to be submerged in soil.''}, ChatGPT generates the explanation {\color{blue}\textit{``The claim is false. According to the information provided, Hughes died in Osterville, Massachusetts, and was interred at Woodlawn Cemetery in the Bronx, New York City.''}} where the explanation actually supports the claim despite it saying the claim is false. Users caught this: \textit{``The explanation sounds like it's actually agreeing with the claim.''} and made the correct judgment. 
    \item In the \texttt{Retrieval + Explanation} condition, users mostly rely on information from retrieved passages ($63.5\%$) and occasionally based on ChatGPT's self-contradiction ($15.9\%$), e.g., users responded \textit{``I made the judgment by summarizing the highlighted areas in the passages.''}~\footnote{We implemented keyword highlighting on the interface as shown in Figure~\ref{fig:interface}.} 
\end{itemize*}
The remaining less common cases are mostly based on personal knowledge or guesses.  


\paragraph{(2) \textbf{When do users make wrong judgments even when the explanations give correct answers?}}
\begin{itemize*}
    \item In \texttt{Explanation}, most users rely on personal knowledge or guess ($46.4\%$), although sometimes explanations have wrong or poor-quality reasoning ($25.0\%$). For example, one user reported \textit{``I once took a tour of Alcatraz and I believe I remember this as being true.''} to the claim \textit{``Within Alcatraz was a music room where inmates could be rewarded for positive behaviors with playing time.''} which is false. 


    \item  In the \texttt{Retrieval + Explanation} condition, users mostly misinterpreted evidence ($38.1\%$) or there was just insufficient evidence ($28.6\%$) and they had to guess.  For example, given the false claim \textit{``The Bee Gees went on tour eleven times.''}, one user responded: \textit{``The articles mention mostly their songs and a couple tours.  I didn't find anything about 11 tours, just a couple of them.  It was mainly songs I saw.''} and they judged the claim to be true even though the ChatGPT explanation is correct: {\color{blue}\textit{``Based on the evidence from Wikipedia, the claim that the Bee Gees went on tour eleven times is not supported. The evidence mentions several tours that the Bee Gees went on, including the 2 Years On Tour, Trafalgar Tour, Mr. Natural Tour, Main Course Tour, Children of the World Tour, Spirits Having Flown Tour, One for All World Tour, High Civilization World Tour, and One Night Only World Tour. However, the evidence does not provide information about any additional tours beyond these mentioned. Therefore, it cannot be concluded that the Bee Gees went on tour eleven times based on the evidence provided.''}}
\end{itemize*}








\section{Conclusion}

Our human studies highlight the false promise of using natural language explanation produced by ChatGPT to aid human fact-checking. 
Humans over-rely on explanations even when they are wrong, making human accuracy worse than showing retrieval or the baseline of not showing any evidence. 
Contrastive explanation mitigates users' over-reliance on wrong explanations, but combining retrieval and explanation does not help further.
Overall, neither of these two approaches significantly beats the retrieval baseline, 
highlighting the need for better methods (\textit{e.g.}, adaptive or customized explanations) for combating over-reliance.


\section*{Limitations and Future Work}

We acknowledge several limitations of this work: 

\noindent
(1) Our experiments are at a limited scale with participants recruited from Prolific. It is possible that other factors such as knowledge of the topics, familiarity with language models, and trust in automation in general, could impact our conclusions and future work should consider scaling up the study with diverse populations to capture such nuances. 

\noindent
(2) We experimented with a limited set of explanation methods and our explanations are all static (\textit{i.e.}, not personalized for different participants). 
Future work could explore how to customize the best sets of evidence for different users in different conditions~\cite{Feng2022LearningTE,Bansal2020DoesTW}. 

\noindent
(3) We observed little benefit from combining retrieval and explanation, future work could further explore 
how to strategically combine retrieval and explanation so that they can potentially complement each other in both accuracy and efficiency. For instance, when the explanation is likely to be correct, we can show users the explanation; but when the explanation is likely to be wrong, we should prioritize showing users the raw retrieved passages. 
This might also require better uncertainty estimation or calibration to help users identify AI mistakes.

\noindent
(4) We used the OpenAI API for generating explanations. All ChatGPT generations were done during July-August 2023, and the specific model checkpoint used is \texttt{GPT-3.5-turbo-0613}. We acknowledge that OpenAI updates its model checkpoints periodically so some results might change in newer versions of the API.

\section*{Ethical Considerations}

In our human studies, we made sure to compensate all participants fairly, with a minimum rate of \$14 per hour. We do not expect any potential mental stress or harm to the participants from the experiments. 
Our work highlights and explores solutions for combatting human over-reliance on AI, which has important societal implications given that LLMs like ChatGPT are being widely used. We hope our results can contribute positively to society by reducing catastrophic harms caused by such over-reliance and also offering practical guidance for how to effectively verify potential fake information on the Internet. 

\section*{Acknowledgement}
We thank Nelson Liu, Xi Ye, Brihi Joshi, Vishakh Padmakumar, Alison Smith-Renner, Helena Vasconcelos, Ana Marasović, Omar Shaikh, and Tianyu Gao for their helpful discussion. 
We also appreciate the feedback from members of the UMD CLIP lab, especially Sweta Agrawal, Pranav Goel, Alexander Hoyle, Neha Srikanth, Rupak Sarkar, Marianna Martindale, Sathvik Nair, Abhilasha Sancheti, and HyoJung Han. 
Chen Zhao is supported by Shanghai Frontiers Science Center of Artificial Intelligence and Deep Learning, NYU Shanghai and Navita Goyal by U.S. Army Grant No. W911NF2120076. 
This material is based on work supported in part by the Institute for Trustworthy AI in Law and Society (TRAILS), which is supported by the National Science Foundation under Award No. 2229885
Any opinions, findings, and conclusions or recommendations expressed in this material are those of the author(s) and do not necessarily reflect the views of the National Science Foundation or the National Institute of Standards and Technology.



\bibliography{anthology,custom}


\appendix


\newpage

\section{Appendix}
\label{sec:appendix}

\subsection{Interface Design}
\label{sec:interface}



\begin{figure*}[h]
\centering
  \includegraphics[width=0.9\textwidth]{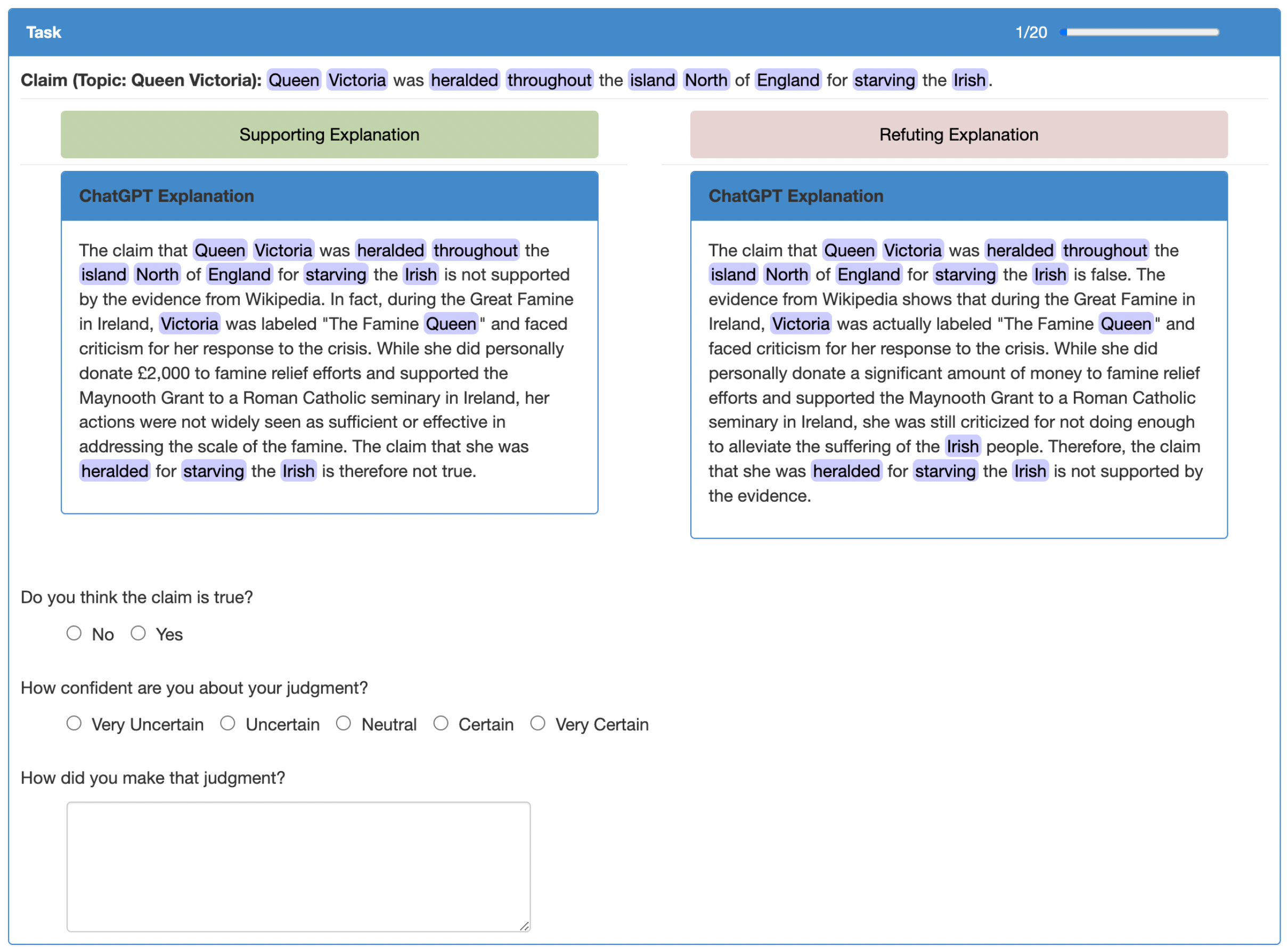}
  \caption{Interface for the contrastive explanation condition. We present ChatGPT's explanations for both sides together to encourage more careful thinking. We also highlight all the keywords to ease reading.}
  \label{fig:interface}
\end{figure*}

\autoref{fig:interface} shows an example user interface for the \texttt{Contrastive Explanation} condition. 
We identify keywords as the non-stopwords in the claim and highlight them in the claims and explanations to aid reading (we also do keyword highlighting in the retrieved passages in the retrieval conditions).  
For the retrieved paragraphs, we rank them by relevance and only show the first paragraph in full by default and annotators can click to expand the other paragraphs.

In the task instructions, we explicitly discourage participants from searching the claims on the internet. 
Each participant verifies $20$ claims one by one. We provide a tutorial at the beginning of the study. We include two attention check questions at different points in the study asking participants' selection from the most recent claim and rejecting the responses from users who fail both attention checks.

\subsection{Additional Plots}
\label{sec:additional_plots}

\begin{figure*}[ht]
    \centering
         \centering
         \includegraphics[width=0.88\textwidth]{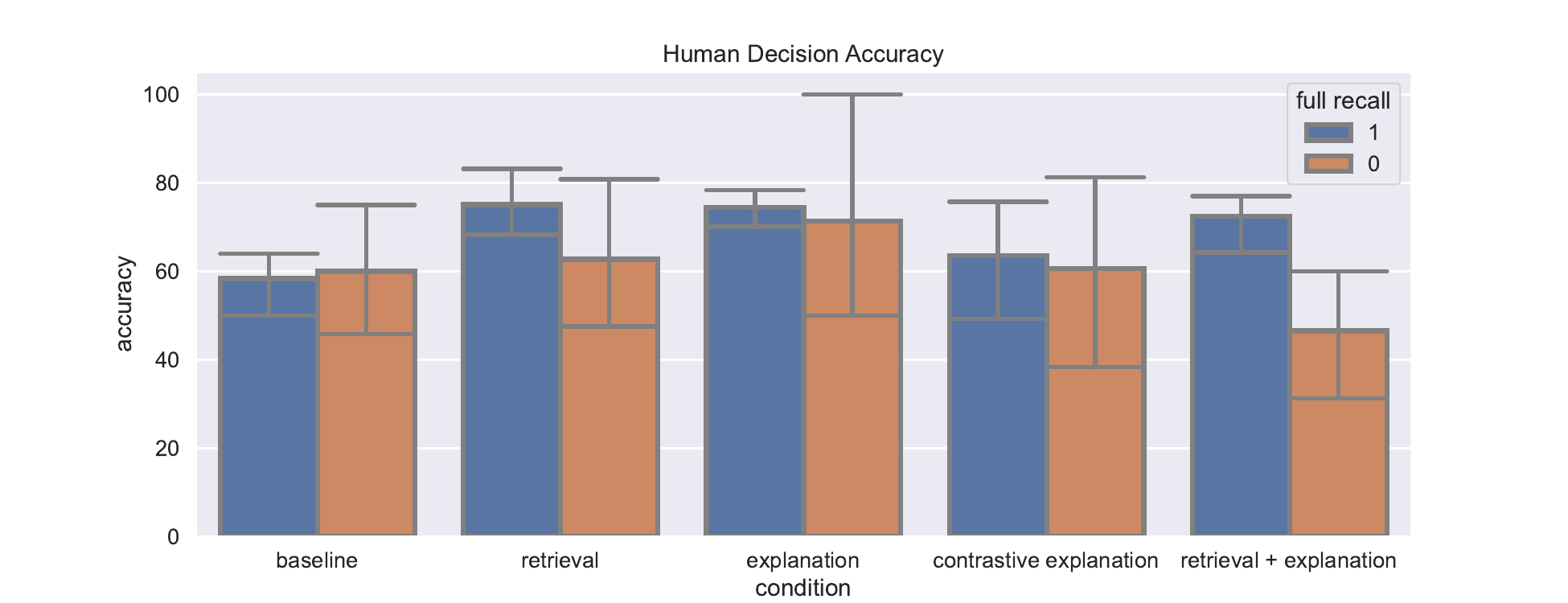}
    \caption{
    Human accuracy broken down by retrieval recall. Human accuracy is lower when the retrieval recall is low. 
}    
\label{fig:analysis_recall}
\end{figure*}

\begin{figure*}[ht]
    \centering
         \centering
         \includegraphics[width=0.8\textwidth]{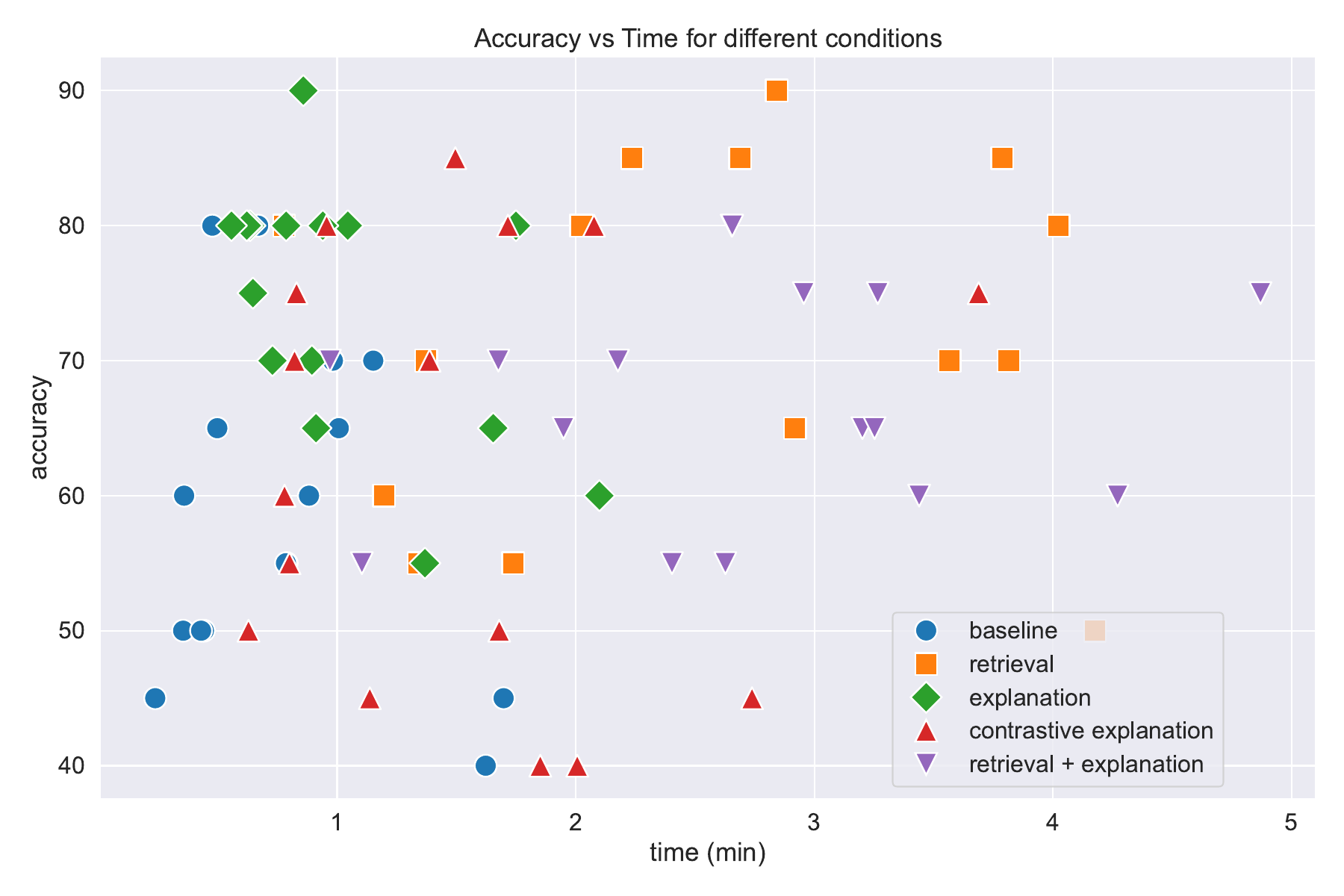}
    \caption{
    Correlation between each participant's average decision accuracy (y-axis) and time (x-axis). We do not observe a strong correlation overall. 
}    
\label{fig:acc_time_corr}
\end{figure*}

\begin{figure*}[ht]
    \centering
     \begin{subfigure}[b]{0.46\textwidth}
         \centering
         \includegraphics[width=\textwidth]{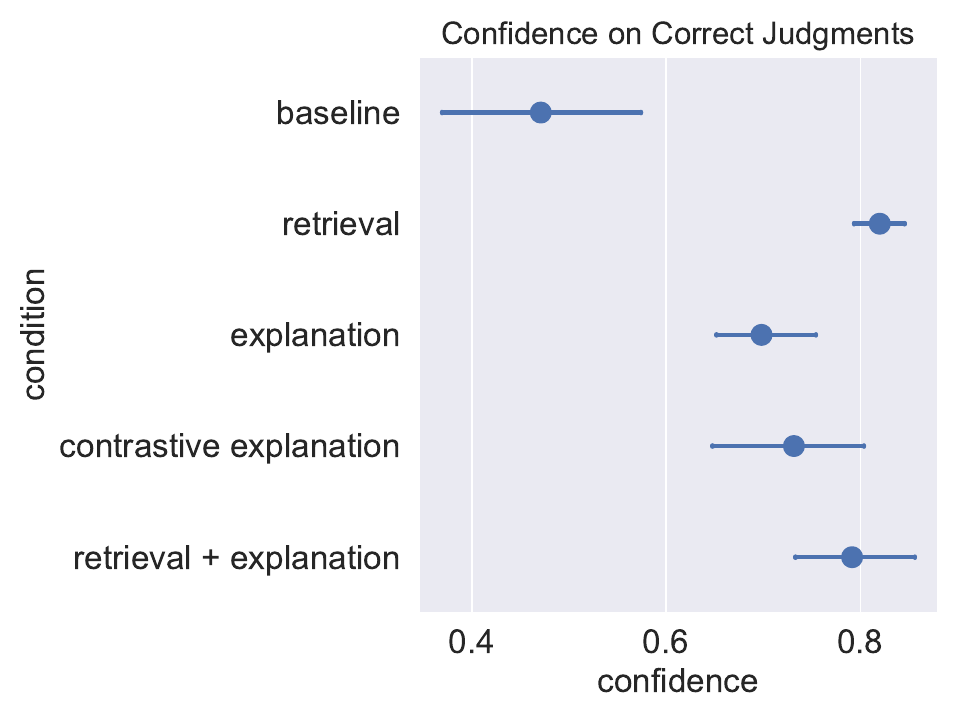}
         \caption{Human confidence in their correct judgments.}
         \label{fig:experiment_2_confidence_correct}
     \end{subfigure}
     \hfill
     \begin{subfigure}[b]{0.46\textwidth}
         \centering
         \includegraphics[width=\textwidth]{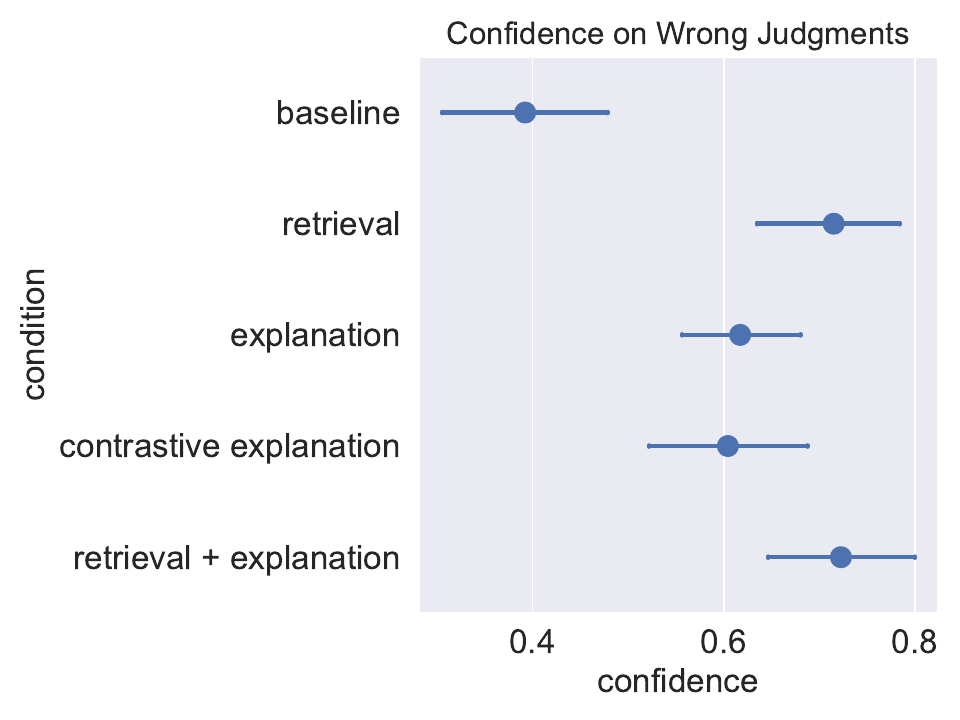}
         \caption{Human confidence in their wrong judgments.}
         \label{fig:experiment_2_confidence_wrong}
     \end{subfigure}
    \caption{
    Human confidence broken down by their correct and wrong judgments. Users are over-confident are wrong judgments. 
}    
\label{fig:experiment_2_confidence}
\end{figure*}

We present several additional plots: Figure~\ref{fig:analysis_recall} plots the human decision accuracy broken down by retrieval recall; Figure~\ref{fig:acc_time_corr} plots the correlation between each participant's average decision accuracy and time used; Figure~\ref{fig:experiment_2_confidence} plots the human confidence broken down by their correct and wrong judgments.

\subsection{Additional Related Work}
\label{sec:more_related_work}

\textbf{On Fact-Checking}: Fact-checking is a well-established task in NLP
where the typical task format is to input evidence text (\textit{e.g.}, retrieved from Wikipedia) and the claim to the model and output a label of support or refute (or sometimes a third class of not enough information)
~\cite{Vlachos2014FactCT,Thorne2018AutomatedFC}. 
Automated fact-checking systems often make use of multiple evidence pieces for making predictions, and optionally generating generations along with predictions~\cite{Popat2018CredEyeAC,Popat2018DeClarEDF,Chen2023ComplexCV}. Various HCI studies have also analyzed how fact-checking systems are used by domain experts such as journalists~\cite{FloresSaviaga2022DatavoidantAA,Beers2019ExaminingTD}.

\noindent
\textbf{On Explanations}:
Explanations have been long sought as a useful tool to help users, not only in understanding AI predictions~\cite{Lee2004Trust} but also aiding them in calibrating their reliance on these predictions \cite{Bussone2015Role}.  
Some works find that explanations can support human-AI decision-making by exceeding both human-alone or AI-alone performance~\cite{Feng2018WhatCA, bowman2022measuring}, whereas some other works find that explanations lead to worse human-AI performance~\cite{Alufaisan2021Does,Bansal2020DoesTW,Ming2021AreHelpful}. 
\citet{Vasconcelos2022ExplanationsCR} and \citet{Fok2023Verifiability}  argue that to facilitate complementary human-AI decision-making, explanations must aid users in verifying the AI prediction to yield truly complementary human-AI performance. 
Explanations targeting verifiability have indeed shown promising avenues in human-AI collaborations~\cite{Feng2018WhatCA, Vasconcelos2022ExplanationsCR, Goyal2023Background}.

\noindent
\textbf{On Explanations for Mitigating Over-Reliance}: 
In line with explanations, model indicators such as confidence~\cite{Zhang2020EffectOC} and accuracy~\cite{Yin2019UnderstandingTE} have been found to yield mixed benefits. On the one hand, uncertainty indicators can promote slow thinking~\cite{Prabhudesai2023UnderstandingUH}, helping users calibrate trust in AI prediction. On the other hand, humans find it difficult to interpret numbers, leading to limited utility of such indicators~\citet{Zhang2020EffectOC}. Further, these indicators can be unreliable as models' accuracy in-the-wild may differ from the reported accuracy on the evaluation set \cite{Chiang2021YouBS} and models' confidence tend to be uncalibrated \cite{Guo2017OnCO}. 
To resolve these limitations, 
\citet{Bussone2015Role} find that detailed explanations exacerbates the over-reliance on the model predictions, whereas less detailed explanations lead to distrust in the model, but increases users' self-reliance. 
\citet{Miller2017ExplanationIA} provides theoretical groundings for why contrastive explanations are naturally human and thus could be more effective. Our experiments provide nuanced empirical findings that contrastive explanations reduce overreliance at the cost of lower accuracy on cases where non-contrastive explanations are originally correct.

\noindent
\textbf{Retrieval for Factuality and Verification}:
Retrieval augmentation (or retrieval-augmented generation; RAG) involves retrieving relevant evidence from a trustworthy external corpus to ground the language model's generation~\cite{Lewis2020RetrievalAugmentedGF,Gao2023RetrievalAugmentedGF}. Retrieval augmentation has been shown to complement parametric knowledge of language models and reduce hallucination~\cite{Mallen2022WhenNT}. Throughout the paper, we adopt the conventional RAG pipeline to ground all explanation generation on retrieved evidence from Wikipedia to improve the factuality of the explanation. On the other hand, retrieved evidence can effectively aid human verification of factual correctness of language model generations~\cite{Min2023FActScoreFA}, which we adopt as a baseline to compare with human verification assisted by natural language explanation generated by models.

\end{document}